\documentclass[letterpaper, 10 pt, conference]{ieeeconf}  

\IEEEoverridecommandlockouts                              

\usepackage{cite}
\usepackage{amsmath,amssymb,amsfonts}
\usepackage{algorithmic}
\usepackage{graphicx}
\usepackage{textcomp}
\usepackage{xcolor}
\usepackage{amsmath}
\usepackage{comment}
\usepackage{gensymb}
\usepackage{breqn}
\usepackage{orcidlink}

\title{\LARGE \bf
Bilinear Model Predictive Control Framework of the OncoReach, a Tendon-Driven Steerable Stylet for Brachytherapy}

\author{Pejman Kheradmand\orcidlink{0000-0001-8579-1443}$^{1}$, Behnam Moradkhani\orcidlink{0009-0004-1609-2779}$^{1}$, Mir Masoud Ale Ali\orcidlink{0009-0003-0752-3568}$^{1}$, \\ Keith Sowards\orcidlink{0009-0000-0852-1976}$^{2}$, Scott R. Silva\orcidlink{0000-0002-6028-3685}$^{2}$ and Yash Chitalia\orcidlink{0000-0002-6291-3492}$^{1}$      
\thanks{$^{1}$P. Kheradmand, B. Moradkhani, M.M. Ale Ali, and Y. Chitalia are with the Healthcare Robotics and Telesurgery Laboratory (Heartlab), University of Louisville, Louisville, KY, USA.}
\thanks{$^2$ K. Sowards and S. R. Silva are with the Department of Radiation Oncology, University of Louisville School of Medicine, Louisville, KY, USA.}
\thanks{\textit{P. Kheradmand and B. Moradkhani have contributed equally to this work.}}
\thanks{Corresponding Author: Pejman Kheradmand ({\tt \small pejman.kheradmand@louisville.edu})}
}

\begin{document}
\maketitle
\thispagestyle{empty}
\pagestyle{empty}

\begin{abstract}
Steerable needles have the potential to improve interstitial brachytherapy by enabling curved trajectories that avoid sensitive anatomical structures. However, existing modeling and control approaches are primarily developed for custom needle designs and are not directly applicable to stylets compatible with commercially available clinical needles. This paper presents a bilinear model predictive control (MPC) framework for a tendon-driven steerable stylet integrated with a standard brachytherapy needle. A geometric bilinear model is formulated with three virtual inputs (an insertion speed and two bending rates) which are mapped to physically realizable inputs consisting of the insertion speed and the associated tendon tensions. The approach is validated through simulations and physical insertion experiments in tissue-mimicking phantom material using image-based tip tracking. 
While open-loop model validation yielded estimation errors below $2$~mm, corresponding to $3\%$ of the inserted needle length, and closed-loop fixed-target tracking achieved an error as low as $1.45$~mm, corresponding to $1.7\%$ of the inserted length, experiments showed larger position errors in certain bending directions, reaching $8.3$~mm, or $7.8\%$ of the inserted length.
Overall, the results demonstrate the feasibility of fixed-target positioning and moving-target trajectory tracking for clinically compatible steerable brachytherapy systems, while highlighting necessary areas for future improvements in calibration and sensing.
\end{abstract}

\section{Introduction}
High-dose rate (HDR) interstitial brachytherapy (ISBT) is a standard-of-care procedure for treating advanced malignancies, such as locally advanced cervical and prostate cancers~\cite{Chargari2019, CervicalCancerVersion32019}. 
Globally, prostate cancer is the most frequently diagnosed cancer in men in over half the countries of the world, with an estimated 313,780 new cases and 35,770 deaths expected in the United States alone in 2025~\cite{CancerStatistics2025, GlobalCancerStatistics2020}.
Similarly, cervical cancer remains a leading cause of cancer mortality in women globally, with 604,000 new cases and 342,000 deaths reported worldwide in 2020, and concerning trends show incidence is rising among young women in the United States~\cite{CervicalCancerUSWomen, GlobalCancerStatistics2020}.
For aggressive cases, such as Gleason score 9-10 prostate cancer,  brachytherapy improves cancer-specific survival~\cite{RadicalProstatectomy}.
The fundamental goal of brachytherapy is to achieve highly conformal radiation therapy while minimizing exposure to organs at risk (OARs)~\cite{Chargari2019}.
However, the complex internal anatomy, deep-seated tumors, and proximity of sensitive healthy tissues make precise needle placement a formidable clinical challenge~\cite{Okazawa_HandHeld}.
Current brachytherapy procedures rely on the manual insertion of rigid, straight needles, restricting surgical planning entirely to linear paths.
This limitation frequently results in suboptimal positioning, as straight needles cannot bypass anatomical obstacles or reach asymmetrical tumor extensions, decreasing the intended target dose.
\begin{figure}
\begin{center}
\centerline{\includegraphics[width=0.85\linewidth,keepaspectratio]{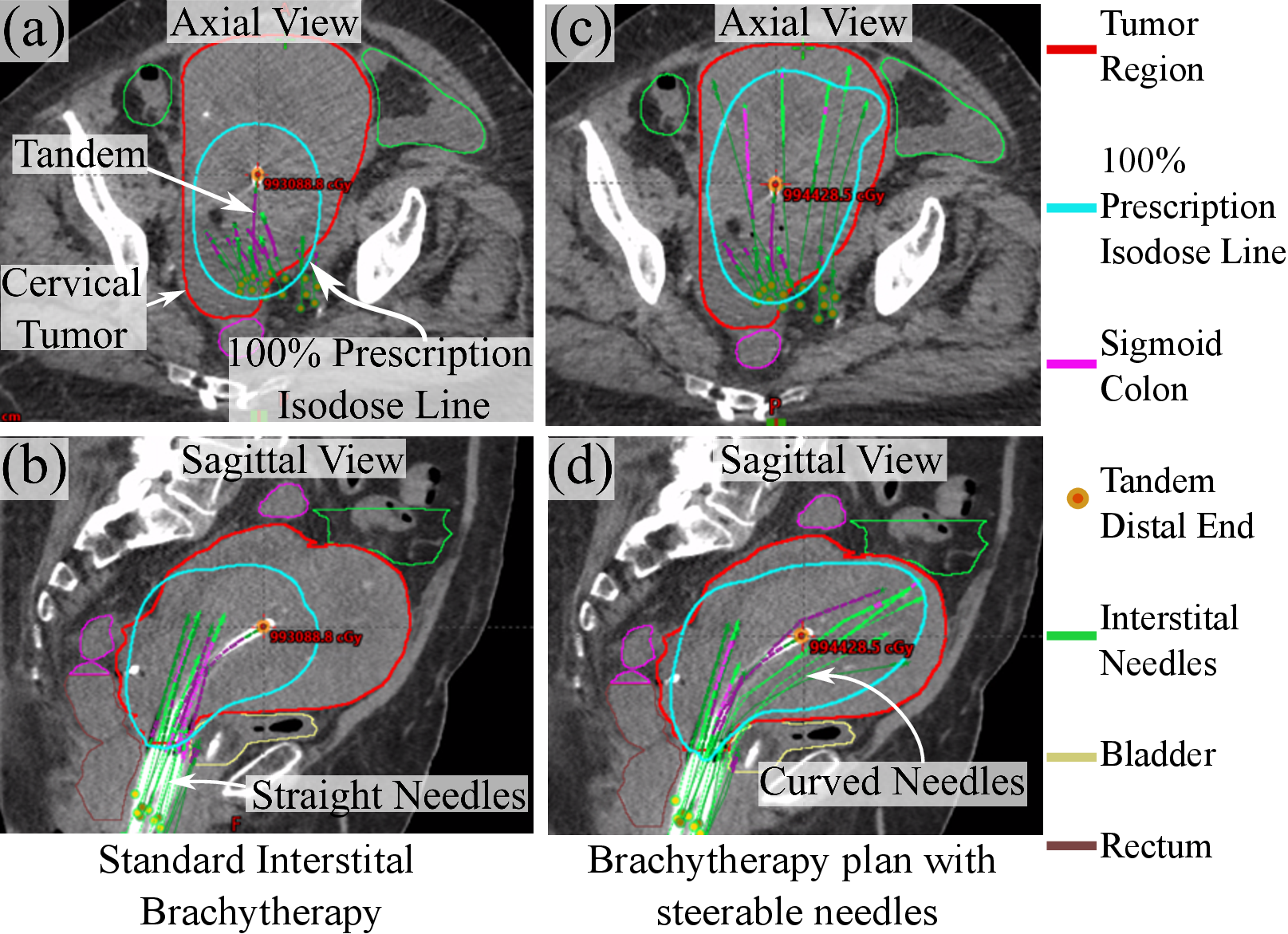}}
\caption{Computed tomography (CT) images of a patient with medically inoperable carcinosarcoma involving the cervix and entire uterus treated with brachytherapy. The tumor is outlined in red. (a) Axial view and (b) sagittal view of standard interstitial brachytherapy (ISBT) using straight needles, where the $100\%$ prescription isodose line (cyan) covers only the distal inferior uterus and cervix due to anatomical limitations. (c) Axial view and (d) sagittal view of a hypothetical ISBT plan using steerable needles, enabling safe dose delivery to the distal uterine fundus while limiting exposure to organs at risk, including the rectum (brown), bladder (yellow), sigmoid colon (purple), and small intestine (green).}
\label{fig:MRI}
\end{center}
\vspace{-9 mm}
\end{figure}
A clinical case of a medically inoperable carcinosarcoma involving the cervix and entire uterus is shown in Fig.~\ref{fig:MRI}. The tumor region is outlined in red. Fig.~\ref{fig:MRI}(a, b) presents axial and sagittal CT views of ISBT performed using straight needles guided by a rigid metallic stylet. Due to anatomical constraints, the $100\%$ prescription isodose line (cyan outline) safely covers only the distal inferior uterus and cervix, as further needle advancement risks damage to surrounding organs at risk.
In contrast, Fig.~\ref{fig:MRI}(c, d) illustrates axial and sagittal views of a hypothetical treatment plan using steerable brachytherapy needles. The steerable needles enable safe targeting of the distal fundal region of the uterus while maintaining limited radiation exposure to nearby OARs, including the rectum, bladder, sigmoid colon, and small intestine.
Several studies have introduced steerable needle technologies to address the limitations of conventional straight needles. Donder et al.~\cite{Donder2023} proposed a programmable bevel-tipped needle capable of generating controlled curvature during insertion. However, the proposed needle was intended for single-use applications, which makes the procedure economically impractical for clinical deployment.
Deaton et al.~\cite{DeatonToward2021} and Gunderman et al.~\cite{Gunderman2023} introduced steerable stylet designs that navigate within standard hollow needles. Despite their promise, limited axial rigidity in their designs reduces steering efficiency during insertion into dense tissue. 
Designs introduced by De Vries et al.~\cite{VriesAxiallyrigid}, as well as the Tendon‑Assisted Magnetically Steered (TAMS) robotic stylet~\cite{TAMS2024} and the OncoReach system~\cite{oncoreach2026}, have demonstrated the feasibility of integrating tendon‑driven robotic stylets with commercially available brachytherapy needles. In particular, the latter two systems achieve high bending compliance while maintaining the axial stiffness required for effective tissue traversal, enabling access to lateral or anatomically constrained targets from safer, less invasive insertion paths. Despite these advancements, achieving precise, predictable control of commercially available brachytherapy needles during insertion remains an open challenge.

Previous studies have investigated needle-tissue interaction modeling and path planning for steerable needle insertion~\cite{RoboticZhao2025, ModelDatla2014}. Padasdao et al.~\cite{MechanicsPadasdao2024} proposed a model to predict the trajectory of a tendon-driven needle within multilayer phantom tissue; however, no control strategy was incorporated. Rucker et al.~\cite{SlidingRucker2013} employed a bevel-tipped needle and applied a sliding-mode control law to regulate needle position in \emph{ex vivo} liver tissue, although the approach did not utilize commercially available clinical needles. Similarly, several studies~\cite{KinematicsKarimi2022, ClosedLafreiniere2024, ModelPadasdao2024, DesignBerg2015, SaferBentley2022} relied on custom-designed needle architectures, limiting direct clinical translation. Misra et al.~\cite{NeedleMisra2008, MechanicsMisra2010} developed mechanics-based finite element models to characterize bevel-tipped needle behavior in tissue; however, control design and experimental implementation were not addressed.
Deaton et al.~\cite{NancyTowardsSteering2023} proposed an adaptive controller a steerable stylet compatible with ISBT needles; however, the maximum steering displacement was limited to $4$~mm, which is insufficient for brachytherapy.
Model Predictive Control (MPC) has also been explored for steerable needle applications. Hussain et al.~\cite{ModelHussain2025} proposed an MPC framework combined with hierarchical supervisory logic and demonstrated it in simulation using a bevel-tipped needle. Hauser et al.~\cite{FeedbackHauser2009} introduced a feedback MPC framework capable of steering a needle along a three-dimensional helical path, although no physical experimental validation was reported. Morley et al.~\cite{SteeringMorley2022} combined recurrent neural networks (RNNs) with MPC, where the RNN learned system dynamics from experimental and simulation data; however, the method was not validated through physical implementation. Collectively, existing studies either focus on modeling without closed-loop control, employ non-clinical custom needle designs, or lack experimental validation in realistic tissue environments, leaving accurate control of clinically compatible steerable stylets during tissue insertion largely unexplored.

To address this gap, this work presents the development and closed-loop control of the OncoReach. The main contributions of this work are summarized as follows:
\begin{itemize}
\item Development of a control framework for steerable stylets that is directly compatible with commercially available ISBT needles.
\item Formulation of a Bilinear MPC framework to regulate the insertion trajectory, together with a virtual-to-physical input mapping that converts desired bending rates and insertion speed into corresponding tendon tensions.
\item Experimental validation through simulations and physical insertion experiments in a realistic tissue-mimicking phantom, demonstrating substantially larger steering displacement compared to prior steerable stylet systems.
\end{itemize}

The remainder of this paper is organized as follows: Section~\ref{sec:ModelControl} presents the needle-tissue interaction dynamics, the Bilinear MPC design, and the input mapping strategy. Section~\ref{sec:Simulations} evaluates the proposed control framework through fixed-target and trajectory-tracking simulations. Section~\ref{sec:DesignAndExperiment} describes the experimental setup and the physical validation of the approach. 
Finally, Section~\ref{sec:conclusion} concludes the paper and discusses limitations and directions for future work.

\section{Model \& Control}\label{sec:ModelControl}
\subsection{Needle-Tissue Interaction Dynamics}
To construct a general geometric model for the trajectory of a steerable needle within a tissue environment, the needle tip position $\textbf{p}(t)\in\mathbb{R}^3$, at time $t$, is first defined as a state variable of the system. To capture the curvature of the needle trajectory, several representations of orientation and directional evolution may be employed, including rotation matrices, Euler angles, or unit direction vectors. In this work, the needle tip direction is represented by a unit direction vector $\hat{\textbf{d}}(t)=(d_x,d_y,d_z)^T$ (with $d_x, d_y, d_z$ being scalars), which directly defines the instantaneous direction of the needle tip trajectory. This choice provides a relatively compact representation of the needle kinematics while remaining well suited for geometric modeling of curved needle motion in tissue.

\begin{figure}
\begin{center}
\centerline{\includegraphics[width=0.80\linewidth,keepaspectratio]{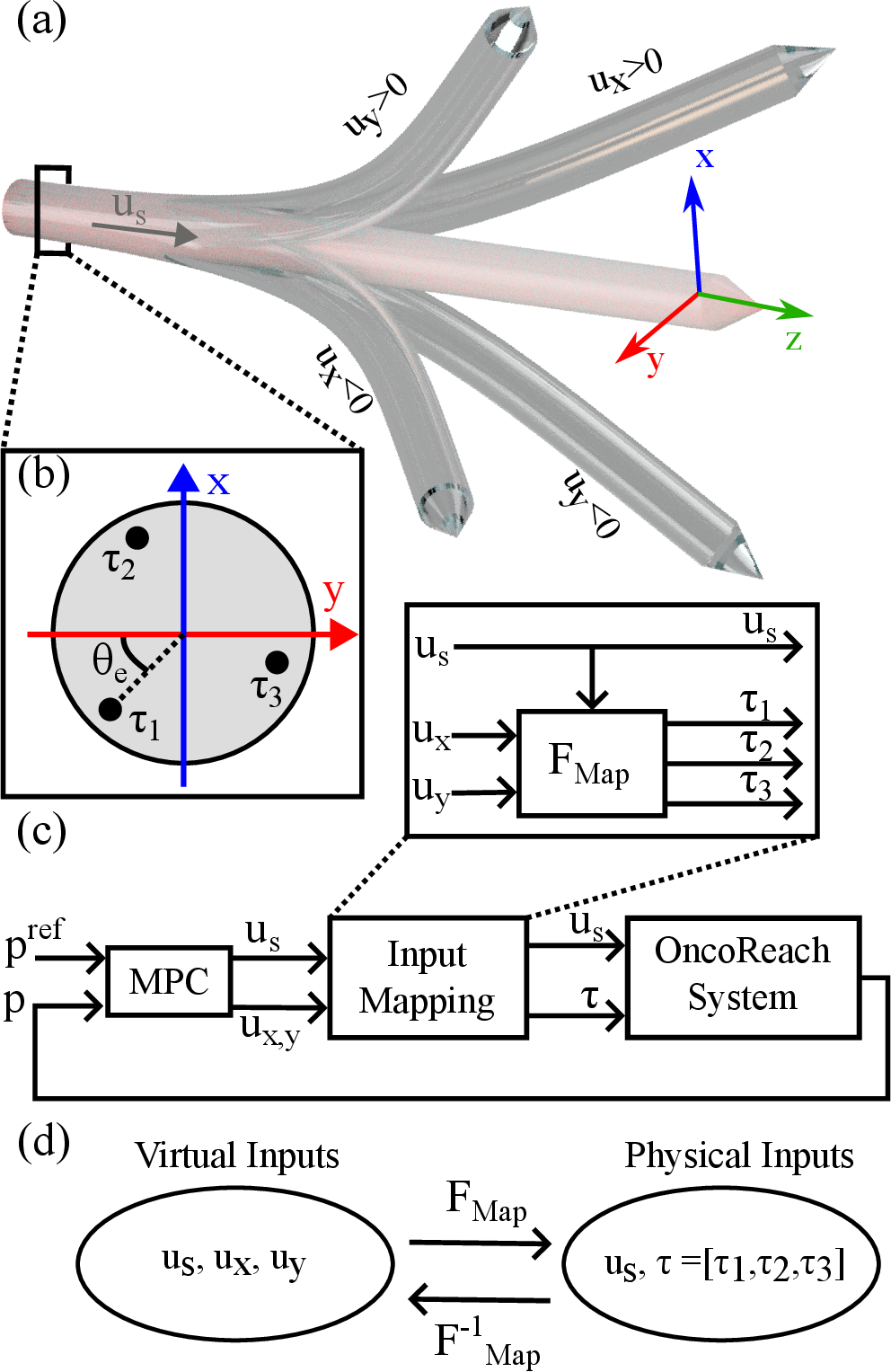}}
\caption{Overview of the proposed modeling and control framework. 
(a) Needle bending in different directions. $u_x$ and $u_y$ denote bending about the $x$- and $y$-axes, respectively. 
(b) Tendon-driven stylet cross-section view showing the orientation of the actuation tendons. 
(c) Bilinear MPC feedback control structure. 
(d) Virtual-to-physical input mapping between bending rates and tendon tensions.}
\label{fig:Model_Control}
\end{center}
\vspace{-5 mm}
\end{figure}

The state equations are constructed based on two intuitive observations regarding needle–tissue interaction dynamics. First, the rate of change (denoted by the dot operator) for $\textbf{p}(t)$ is governed by the insertion speed $u_s(t)$ projected along the instantaneous $\hat{\textbf{d}}(t)$. Second, variations in the trajectory direction arise from bending of the needle tip induced by the stylet actuation. Together, these principles define a general kinematic model for steerable needle motion. In the specific system considered in this work, no base rotation is available as an actuation input, and the actuation tendons are routed through straight channels from base to tip, resulting in negligible tendon-induced twisting. Consequently, twisting effects and associated helical curvature components are neglected, and the model is restricted to bending-induced directional changes.
\begin{equation}
    \dot{\textbf{p}}(t) = u_s(t) \hat{\textbf{d}}(t) \\
\end{equation}
\begin{equation}
    \dot{\hat{\textbf{d}}}(t) = \hat{\textbf{d}}(t) \times 
    \begin{pmatrix}
    u_x(t) \\
    u_y(t) \\
    0
    \end{pmatrix}.
\end{equation}
Note that $u_x(t)$ and $u_y(t)$ denote virtual inputs governing the rate of directional change of the needle tip in the horizontal and vertical planes, respectively (See Fig. \ref{fig:Model_Control}(a) and (b)). Physically, $u_x(t)$ and $u_y(t)$ depend on the actuation tendon tensions $\boldsymbol{\tau}(t)$ and the insertion speed $u_s(t)$. At this stage, however, $u_s$, $u_x$, and $u_y$ are treated as independent control inputs, referred to as ``virtual inputs'', to facilitate a simpler and more general formulation of the model~\cite{WebsterNonholonomic}. Under this representation, the resulting state equations can be expanded element-wise into a purely bilinear form, consisting exclusively of terms that are products of state variables and control inputs. For notational compactness, explicit time dependence is omitted from the state and input variables in the remainder of this paper.
\begin{equation}
\begin{split}
    \dot{x}=u_s d_x \hspace{25pt} , \hspace{25pt} \dot{y}=u_s d_y \hspace{25pt} , \hspace{25pt} \dot{z}=u_s d_z \\
    \dot{d_x}=-d_z u_y \hspace{10pt} , \hspace{10pt} \dot{d_y}= d_z u_x \hspace{10pt} , \hspace{10pt} \dot{d_z}= d_x u_y - d_y u_x.
\end{split}
\end{equation}
Introducing state vector $\textbf{s}=(\textbf{p},\hat{\textbf{d}})^T$, the overall bilinear virtual system is represented as:
\begin{equation}
    \dot{\textbf{s}}=u_s \textbf{B}_1 \textbf{s}+u_x \textbf{B}_2 \textbf{s}+u_y \textbf{B}_3 \textbf{s}
\end{equation}
where $\textbf{B}_1$, $\textbf{B}_2$, and $\textbf{B}_3$ are constant system matrices.
\begin{equation}
    \textbf{B}_1=
    \begin{bmatrix}
        \textbf{0} & I \\ \textbf{0} & \textbf{0}
    \end{bmatrix} ;
    \textbf{B}_2=
    \begin{bmatrix}
        \textbf{0} & \textbf{0} \\ \textbf{0} & G
    \end{bmatrix};
    \textbf{B}_3=
    \begin{bmatrix}
        \textbf{0} & \textbf{0} \\ \textbf{0} & H
    \end{bmatrix}.
\end{equation}
Note that $\textbf{0}\in \mathbb{R}^{3\times3}$ represents zero square matrix. Additionally, $\textbf{G}$ and $\textbf{H}$ (both $\in \mathbb{R}^{3\times3}$) are used as auxiliary matrices, used for compact representation.
\begin{equation}
        \textbf{G}=
    \begin{bmatrix}
        0 & 0 & 0 \\ 0 & 0 & 1 \\ 0 & -1 & 0
    \end{bmatrix};
        \textbf{H}=
    \begin{bmatrix}
        0 & 0 & -1 \\ 0 & 0 & 0 \\ 1 & 0 & 0
    \end{bmatrix}.
\end{equation}
\subsection{Bilinear MPC Design}
Given that the resulting virtual system is described by a purely bilinear state-space model, a bilinear MPC can be naturally employed for needle trajectory tracking. Compared to nonlinear MPC (NMPC), the bilinear structure avoids general nonlinear dynamics, leading to reduced computational complexity and improved suitability for real-time implementation. To enable the design and implementation of the bilinear MPC, the continuous-time system is first discretized.
\begin{equation}
    s_{k+1} = [\textbf{I}+T_s(u_s \textbf{B}_1 + u_x \textbf{B}_2 + u_y \textbf{B}_3)]s_k
\end{equation}
where $s_{k}$ is the current state (state at the current step $k$), $s_{k+1}$ is the next state, $\textbf{I}\in \mathbb{R}^{3\times3}$ depicts the identity matrix, and system time step is depicted by $T_s$. MPC operates by optimizing a predefined cost function over a finite window of predicted future system states, commonly referred to as the control horizon. The cost function typically comprises two components: a system output tracking term that penalizes deviations between the predicted output and the desired reference output, and an input regulation term that penalizes excessive control effort. This formulation enables MPC to balance trajectory tracking performance against actuation smoothness and feasibility while explicitly accounting for system dynamics.
\begin{equation}
    J=\sum_{i=0}^{N} \textbf{e}_{k+i}^T \textbf{Q} \textbf{e}_{k+i} + \textbf{u}_{k+i}^T \textbf{R} \textbf{u}_{k+i}
\end{equation}
where $J$ is the cost function value, $N$ is the control horizon size, $i$ denotes the discrete time step within the control horizon, $\textbf{e}_{k+i}=\textbf{p}_{k+i}-\textbf{p}^{ref}_{k+i}$ is the output position error vector (direction state vector is not considered as an output of the system), $\textbf{Q}$ is the position error weight matrix, and $\textbf{R}$ is the input weight matrix. Both $\textbf{Q}$ and $\textbf{R}$ are diagonal, where each diagonal element specifies the relative importance of penalizing the corresponding state deviation and control input, respectively. This structure allows for intuitive tuning of the controller by independently adjusting the influence of individual states and inputs within the MPC cost function.

\begin{figure}
\begin{center}
\centerline{\includegraphics[width=\linewidth,keepaspectratio]{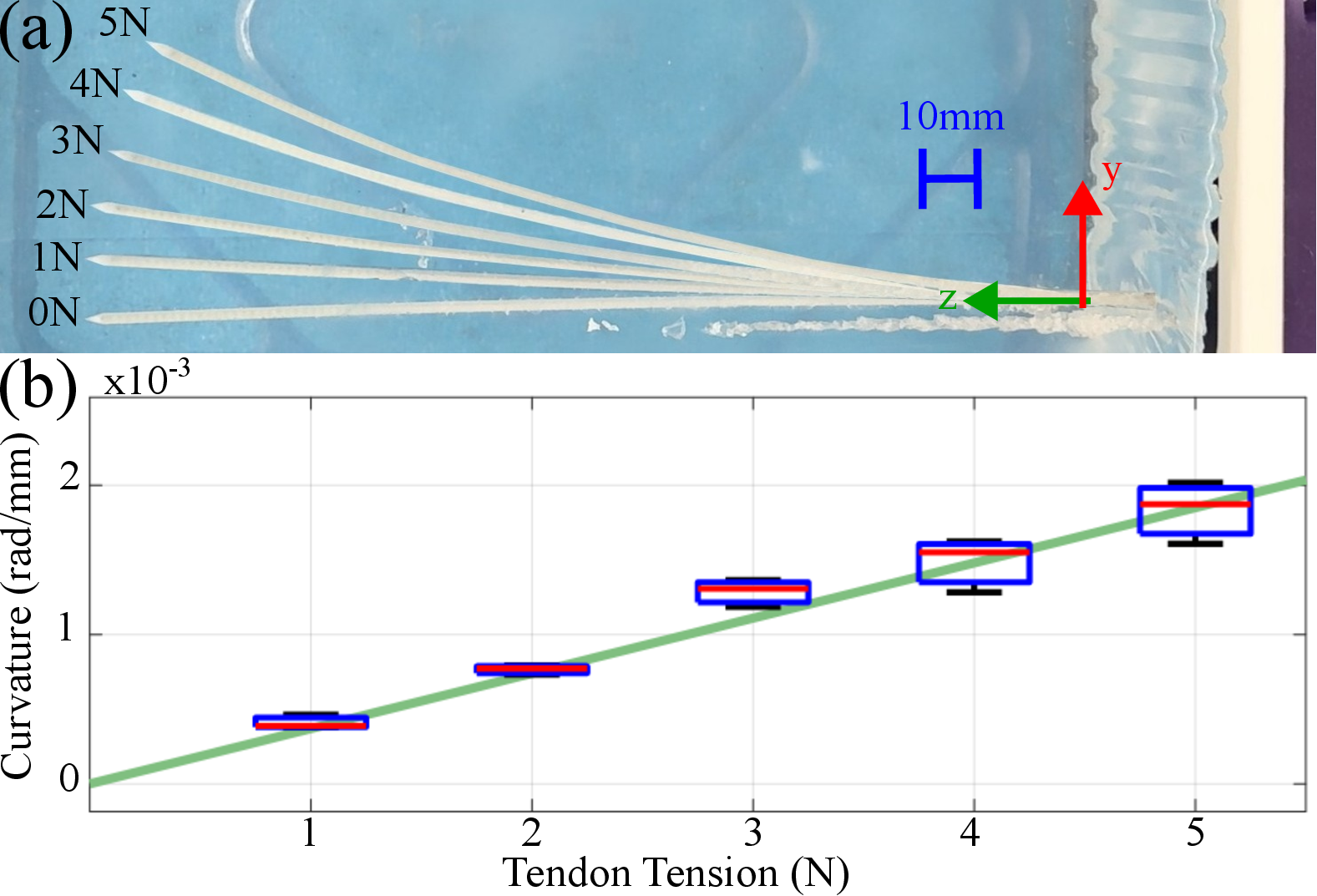}}
\caption{Experimental calibration of the curvature-tension relationship. (a) Needle bending inside phantom tissue under different applied tendon tensions. (b) Measured trajectory curvatures as a function of tendon tension; bar plots represent experimentally estimated curvatures with a zero-offset linear fit.}
\label{fig:Model_Calibration}
\end{center}
\vspace{-8 mm}
\end{figure}

With the cost function defined, the control problem is formulated as the minimization of the objective function $J$ over the control horizon. This optimization problem is solved at each sampling instant, yielding an optimized control input sequence $\textbf{U}_{opt} \in \mathbb{R}^{3\times N}$, which stacks the optimized virtual inputs for the control horizon. 
\begin{equation}
    \textbf{U}_{opt} = \arg \min_{\{u_s, u_x, u_y\}}(J)
\end{equation}
Following the receding-horizon principle, only the first control input in the optimized sequence is applied to the system, while the remaining inputs are discarded. The optimization is then repeated at the next time step using updated measurements. In addition, input constraints can be incorporated into the optimization problem to prevent actuator saturation. Input constraints are particularly important in medical and surgical applications due to safety considerations. In the proposed approach, constraints are enforced on the virtual control inputs employed in the bilinear MPC formulation. Although these inputs will be mapped to physical actuator commands, saturation of the resulting tendon tensions is not explicitly prevented. By contrast, the insertion speed is directly constrained (See Fig. \ref{fig:Model_Control}(c)), as it is common to both the virtual and physical input representations.
\subsection{Input Mapping}
To translate the virtual control inputs generated by the bilinear MPC into physically meaningful actuation commands, a relationship between tendon tensions and needle tip direction rates must be established (See Fig. \ref{fig:Model_Control}(c,d)). This mapping is derived under a set of simplifying yet physically motivated assumptions. In particular, it is assumed that, for a fixed combination of applied tendon tensions, the needle follows a trajectory with constant curvature throughout the insertion. This assumption is commonly adopted in the modeling of steerable needles and forms the foundation of non-holonomic models~\cite{WebsterNonholonomic}. It relies on the premise that the needle shaft is sufficiently flexible to conform to the path dictated by the needle tip without inducing significant additional deformation or tissue damage, allowing the overall trajectory to be governed primarily by the local tip curvature.
\begin{equation} \label{bx and by}
    u_x = \kappa_x(\boldsymbol{\tau})u_s \hspace{10pt} , \hspace{10pt} u_y = \kappa_y(\boldsymbol{\tau})u_s
\end{equation}
where $\kappa_x$ and $\kappa_y$ are needle trajectory curvatures about x- and y-axes and they depend on the actuation tendon tensions, which in our case are $\boldsymbol{\tau}=(\tau_1, \tau_2, \tau_3)^T\in \mathbb{R}^3$.

To determine the curvature components induced in different directions as a function of the applied tendon tensions, the orientation of the tendon channels relative to the needle coordinate frame must first be defined (See Fig. \ref{fig:Model_Control}(b)). Based on this geometric information, two additional simplifying assumptions are introduced to formulate the curvature-tension relationship. First, the resulting needle tip bending and the corresponding trajectory curvature are assumed to obey a superposition principle, whereby the total curvature is obtained as the linear combination of the individual curvature contributions generated by each tendon. Second, the bending behavior associated with each actuation tendon is assumed to be identical, reflecting the symmetric design of the stylet and the equal radial spacing of the tendon channels with respect to the cross-section center. Together, these assumptions enable a compact and tractable mapping between tendon tensions and directional curvature components.
\begin{align}\label{kappax and kappay}
\begin{split} 
     \kappa_x(\boldsymbol{\tau}) = cos(-\theta_e) \kappa(\tau_1) + cos(\frac{2\pi}{3}-\theta_e) \kappa(\tau_2) +
     \\
     cos(\frac{4\pi}{3}-\theta_e) \kappa(\tau_3)
     \\
     \kappa_y(\boldsymbol{\tau}) = sin(-\theta_e) \kappa(\tau_1) + sin(\frac{2\pi}{3}-\theta_e) \kappa(\tau_2) + 
     \\
     sin(\frac{4\pi}{3}-\theta_e) \kappa(\tau_3).
\end{split}
\end{align}
The parameter $\theta_e$ represents the angle between the negative y-axis of the reference coordinate frame and the direction of the first tendon channel (see Fig. \ref{fig:Model_Control}(b)). In addition, $\kappa(.)$ denotes the curvature mapping function that relates the tendon tension values to their corresponding needle tip curvature values. $\kappa(.)$ depends on both the mechanical design of the needle and the properties of the surrounding tissue and therefore cannot be assumed to be universal across different systems or environments. In this work, an experimental calibration procedure is performed to identify an appropriate function definition, which is then used for the implementation of the proposed model and controller.
\begin{figure}
\begin{center}
\centerline{\includegraphics[width=0.8\linewidth,keepaspectratio]{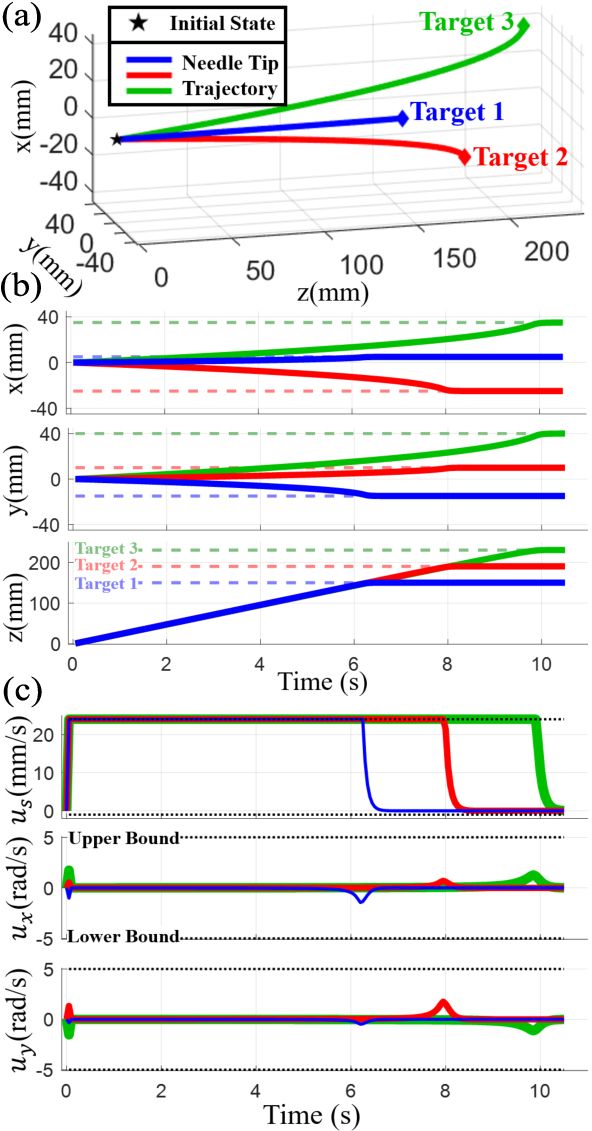}}
\caption{Simulation results for fixed-target scenarios. (a) Needle tip trajectory from the initial position to the target in three-dimensional space. (b) Temporal evolution of the needle tip position, showing convergence to the desired target. (c) Optimal virtual input signals generated by the bilinear MPC.}
\label{fig:Fix_Target_Simulaion}
\end{center}
\vspace{-5 mm}
\end{figure}
\begin{figure*}
\begin{center}
\centerline{\includegraphics[width=0.8\linewidth,keepaspectratio]{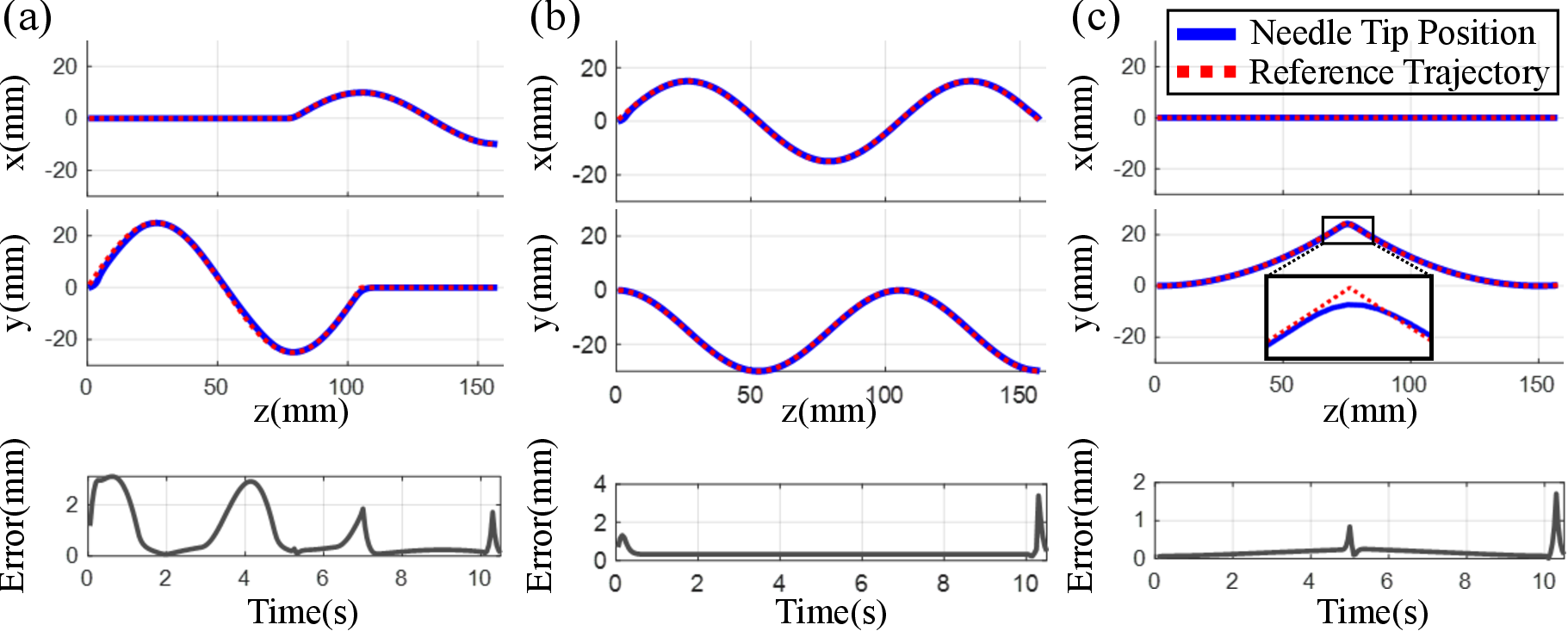}}
\caption{Simulation results for trajectory-tracking scenarios. (a) Arbitrary 3D trajectory tracking, (b) Helical trajectory, and (c) trajectory with a sharp turn.}
\label{fig:Moving_Target_Simulaion}
\end{center}
\vspace{-8 mm}
\end{figure*}
During the calibration experiments, multiple needle insertion trials were performed while applying prescribed tension levels to a selected tendon, resulting in planar bending trajectories.
\textcolor{black}{Although the theoretical model accommodates three-dimensional trajectories, current sensing limitations necessitated the use of a 2D camera-based visualization setup and a color detection algorithm for needle tip tracking (See Fig. \ref{fig:Model_Calibration}(a)).}
For each trial, spatial location data were sampled along the resulting trajectories and subsequently processed to estimate the corresponding curvature values (See the bar plot in Fig. \ref{fig:Model_Calibration}(b)). The collected experimental data exhibit an approximately linear relationship between the applied tendon tension and the resulting curvature (See Fig. \ref{fig:Model_Calibration}(b)).
\begin{equation} \label{kappa tau linear}
    \kappa(\tau_j)= (3.7 \times 10^{-4}) \tau_j \hspace{5pt} ; \hspace{5pt} j=1,2,3.
\end{equation}
While it remains unclear whether this linear trend represents a local approximation of a more general nonlinear relationship or reflects an inherently linear behavior, a detailed investigation of this aspect is deferred to future work. 

Equations (\ref{bx and by}), (\ref{kappax and kappay}), and (\ref{kappa tau linear}) enable the computation of a unique set of bending rates for any given tendon tension combination. However, the implementation of the bilinear MPC requires the inverse of these equations, namely determining the tendon tensions that produce a desired set of bending rates (See Fig. \ref{fig:Model_Control}(c,d)). Deriving this inverse relationship in closed form is not analytically straightforward. Consequently, a numerical approach is adopted to solve the problem. Specifically, a linear least-squares formulation is employed, and the inverse function is computed using the \texttt{lsqlin} solver in MATLAB.



\section{Simulations}\label{sec:Simulations}
Based on the developed model and the proposed bilinear MPC framework, a simulation environment was implemented in MATLAB to evaluate control performance. The MPC optimization problem was solved using the \texttt{fmincon} solver. The simulation parameters were selected as $T_s=0.05$~s, $N=5$, and a total of 210 simulation steps, with the system initialized at $s_0=(0,0,0,0,0,1)^T$. The weighting matrices were chosen as $\textbf{Q}=diag(100,100,200)$ and $\textbf{R}=diag(1,1,1)$ (the weight of position error in z-axis direction was intentionally set higher to avoid overshoots that can possibly puncture vital organs). Virtual input constraints were imposed to reflect practical limits, with bounds $\pm 5 $ rad/s for $u_x$ and $u_y$, and the range $[-1,24]$ m/s for $u_s$.
\subsection{Fixed-Target Simulation}
In the fixed-target simulations, three target points located at arbitrary positions were selected, namely Target~1 $= (5, -15, 150)^T$, Target~2 $= (-25, 10, 190)^T$, and Target~3 $= (35, 40, 230)^T$ (See Fig. \ref{fig:Fix_Target_Simulaion}(a, b)). For each case, the needle tip position and the corresponding control inputs generated by the bilinear MPC were recorded. The simulation results demonstrate that the proposed controller successfully navigates the needle tip to the specified targets, achieving final Euclidean distance errors of $0.144$~mm, $0.137$~mm, and $0.077$~mm for targets 1, 2, and 3, respectively. The control input profiles indicate that the controller primarily exploits the maximum allowable insertion speed $u_s$ to rapidly approach the target region, while the directional inputs $u_x$ and $u_y$ are activated near the final phase of motion to fine-tune the trajectory and regulate convergence toward the target points (See Fig. \ref{fig:Fix_Target_Simulaion}(c)).

\subsection{Trajectory-Tracking Simulation}
In the trajectory-tracking simulations, three distinct reference paths were provided to the control system to evaluate tracking performance under varying levels of geometric complexity. The first trajectory was arbitrarily defined to demonstrate the general tracking capability of the proposed controller and included high-amplitude sinusoidal segments. These segments required aggressive directional virtual inputs $u_x$ and $u_y$, which led to increased Euclidean distance errors at certain time instances, with a maximum error of $3.122$~mm (See Fig. \ref{fig:Moving_Target_Simulaion}(a)). In the second simulation, a helical reference trajectory was selected, and the needle tip closely followed the desired path with minimal tracking error throughout most of the motion. A transient increase in error, reaching $3.4$~mm, was observed at the terminal point of the trajectory, likely due to the abrupt cessation of motion (See Fig. \ref{fig:Moving_Target_Simulaion}(b)). In the third simulation, the controller’s ability to handle sharp corners and sudden directional changes in the reference trajectory was evaluated. The results indicate that the maximum tracking error at the sharp edge remained below $1$~mm, demonstrating smooth input generation and effective regulation of the needle tip in the presence of non-smooth reference trajectories (See Fig. \ref{fig:Moving_Target_Simulaion}(c)).

\section{Experiments and Results}\label{sec:DesignAndExperiment}
\subsection{Experimental Setup}
\begin{figure}
\centering\includegraphics[width=0.85\linewidth,keepaspectratio]{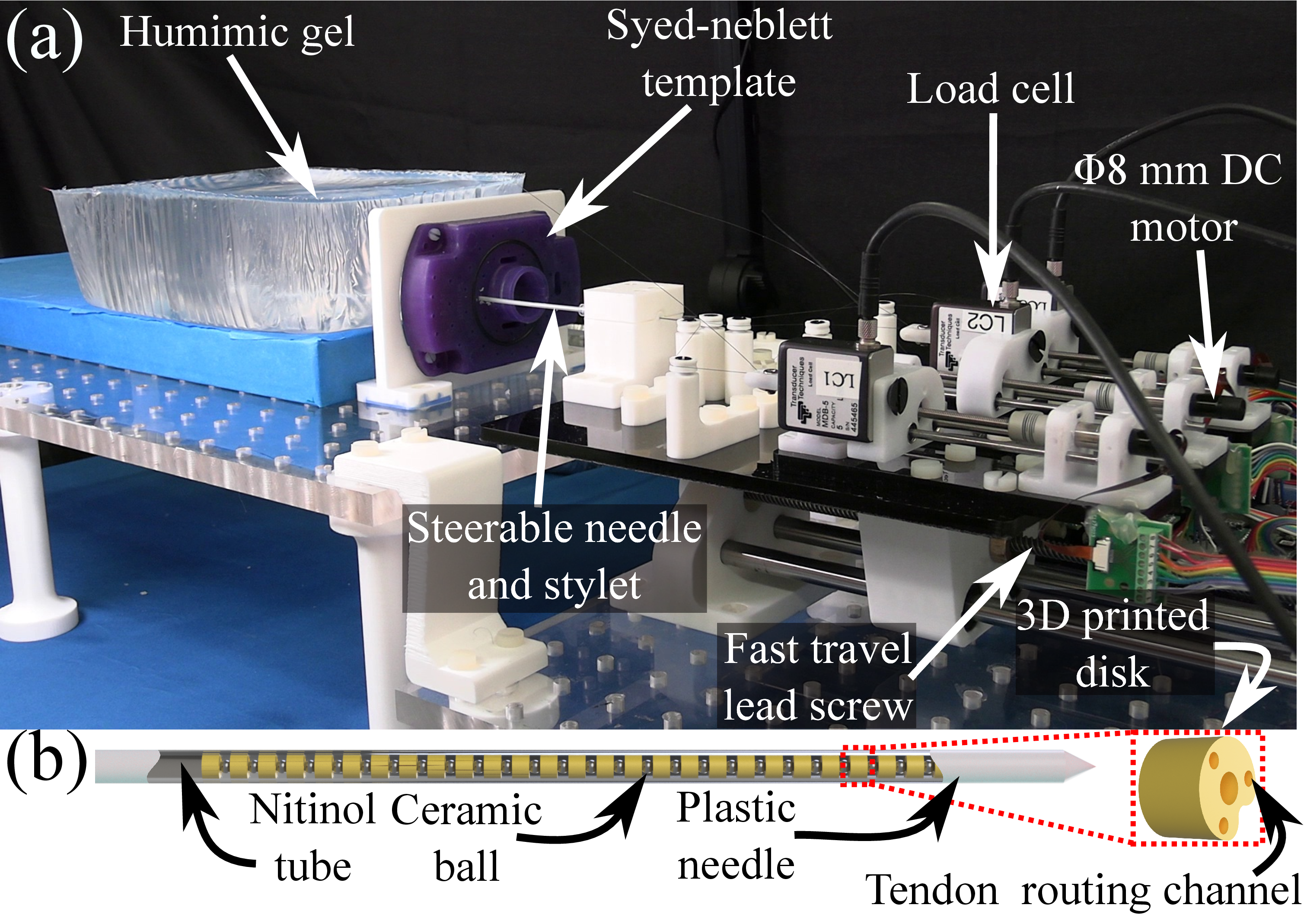}
\caption{Experimental setup. (a) System overview showing the back-end assembly, which includes a DC motor coupled to a fast-travel lead screw to drive the linear insertion motion and the tendon actuation assembly featuring a load cell for tension measurement. (b) The front-end design, illustrating the steerable stylet inserted into the plastic needle. The stylet comprises a nitinol tube integrated with 3D-printed routing disks with tendon routing channels and ceramic spherical joints.}
\label{fig:setup}
\vspace{-5 mm}
\end{figure}
The experimental setup comprises actuation systems designed to govern both the insertion and steering of the stylet. The linear insertion motion of the entire needle assembly is driven by a DC motor (A-max 16 $\phi16$~mm, Precious Metal Brushes CLL, 1.2 Watt) coupled to a fast-travel lead screw. This lead screw is supported by two parallel stainless steel rods to ensure precise and smooth translation of the main stage (see Fig.\ref{fig:setup}(a)). Mounted on top of this translating stage are three independent tendon actuation units. Each unit utilizes a DC motor ($\phi8$~mm, Maxon Metal Brushes, 0.5 Watt) to advance and retract a load cell (MDB-$5$ $5$lb capacity, Transducer Techniques), enabling the system to pull or release the individual steering tendons.
\begin{figure}
\centering\includegraphics[width=0.8\linewidth,keepaspectratio]{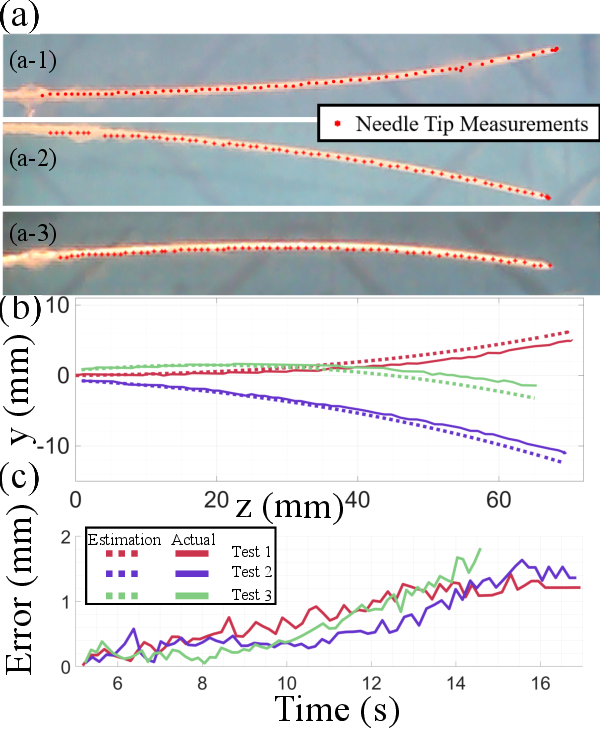}
\caption{Model validation experimental results. (a) Needle trajectories in tissue phantom with needle tip measurements shown for: (a-1) test~1, (a-2) test~2, and (a-3) test~3. (b) Comparison between the actual needle tip trajectories and model estimations. (c) Model estimation errors.}
\label{fig:ModelValidation}
\vspace{-7 mm}
\end{figure}
At the front-end, the steerable stylet is inserted coaxially into a standard ISBT 15-gauge plastic needle. The stylet consists of a proximal nitinol tube integrated with a distal steerable section. This articulated tip is composed of ceramic spherical joints separated by 3D-printed routing disks, which guide the tendons along the length of the stylet (see Fig.\ref{fig:setup}(b)). A Syed–Neblett template (a perforated plastic grid) is positioned before the phantom tissue to guide needle insertion.
By selectively actuating individual tendons via the back-end units, the stylet bends toward the corresponding direction of the applied tendon force.
To evaluate the insertion and steering capabilities of the system under realistic resistive forces, experiments are conducted using a high-fidelity phantom tissue model. The phantom tissue comprises Humimic gel (SimuGel 3, Humimic, Greenville, SC, USA). This material provides a highly accurate simulation environment with a Young's modulus of 0.19~MPa, which has been validated to mimic the properties of human tissue during brachytherapy procedures~\cite{humimic}. By coordinating the linear insertion stage with the tendon actuation units, the needle can be dynamically steered while being inserted into the phantom tissue to navigate toward targeted regions. To obtain real-time needle tip position data during the experiments, a downward-facing webcam was mounted above the phantom tissue, and a MATLAB-based algorithm employing RGB image decomposition was used for tip detection and tracking. Since needle tip tracking relies on image-based detection, planar motion was enforced by setting $u_y = 0$.

\subsection{Model Validation Experiments}

To validate the proposed model, three open-loop experiments were conducted using predefined $u_s$ and $\boldsymbol{\tau}$ inputs. During needle navigation, real-time needle tip position data were recorded. The same input profiles were subsequently applied to the theoretical model, and the estimated tip trajectories were compared with the experimentally measured trajectories (See Fig. \ref{fig:ModelValidation}(a)). The Euclidean distance error between the estimated and experimental tip positions increased with insertion depth. However, in all experiments, the error remained below $2$~mm or $3\%$ of the inserted needle length (See Fig. \ref{fig:ModelValidation}(b)).

\subsection{Experimental Bilinear MPC Implementation}

\begin{figure}
\centering\includegraphics[width=0.80\linewidth,keepaspectratio]{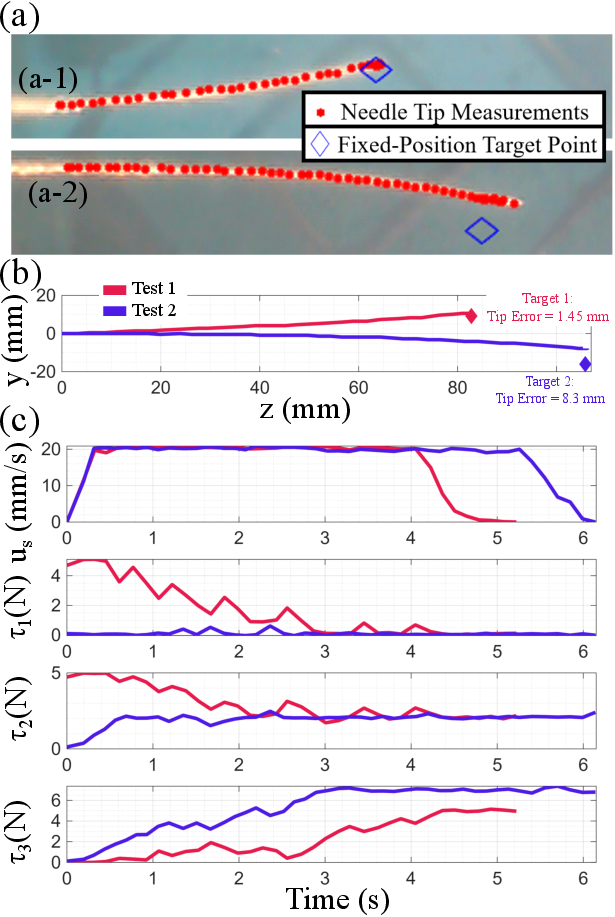}
\caption{Fixed-target experimental results. (a) Needle trajectories in tissue phantom with needle tip measurements and target points shown for: (a-1) test~1 and (a-2) test~2. (b) Needle tip trajectories toward targets requiring opposite bending directions. (c) Corresponding insertion speed and tendon tensions.}
\label{fig:ControlConstantTarget}
\vspace{-7 mm}
\end{figure}

\begin{figure}
\centering\includegraphics[width=0.80\linewidth,keepaspectratio]{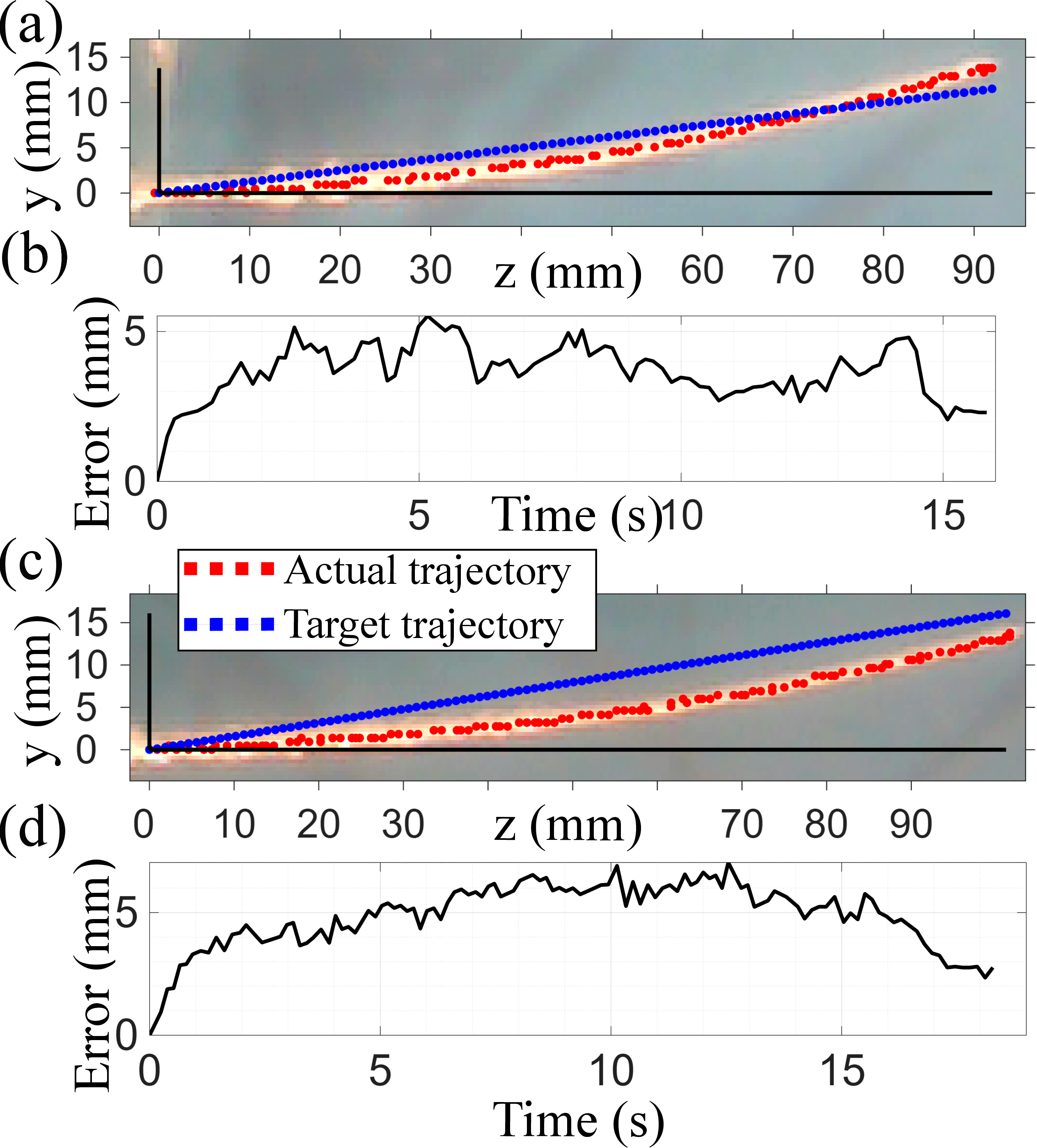}
\caption{Experimental trajectory-tracking results with MATLAB trajectory reconstruction overlaid on the recorded experimental images. (a) Actual and target trajectories for Test 1. (b) Error between the actual and target trajectories for Test 1. (c) Actual and target trajectories for Test 2. (d) Error between the actual and target trajectories for Test 2.}
\label{fig:ControlMovingTarget}
\vspace{-7 mm}
\end{figure}
Ultimately, the bilinear MPC was implemented on the physical needle insertion system to evaluate the practical fixed-target and trajectory-tracking performance of the proposed controller.
Two sets of target points were defined for the fixed-target experiments, as shown in Fig.~\ref{fig:ControlConstantTarget}(a), corresponding to opposite lateral bending directions ($u_x < 0$ and $u_x > 0$). The insertion speed was limited to $20$~mm/s to satisfy actuator constraints, and the resulting insertion speed and tendon tensions are shown in Fig.~\ref{fig:ControlConstantTarget}(b). As discussed previously, the tendon tension values obtained from mapping the virtual bending rates are not guaranteed to lie within admissible bounds. Therefore, as a precautionary measure, a saturation block (with an upper bound of 7~N) is applied to the commanded tendon tensions.

Trajectory-tracking performance was further evaluated through two moving-target trials, where the desired needle tip position was updated continuously based on image measurements. Figure~\ref{fig:ControlMovingTarget} presents the reference trajectories, measured needle tip motion, and corresponding tracking errors for both trials.
Consistent with the proposed simulation studies, the MPC objective function was optimized using the \texttt{fmincon} solver in MATLAB, and the control horizon, initial state, and weighting matrices $\textbf{Q}$ and $\textbf{R}$ were chosen to match those used in simulation. Owing to the computational time required for optimization and image-based tip tracking, the controller sampling time was set to $T_s=1$~s.
While open-loop validation yielded errors below $2$~mm, closed-loop performance varied by direction. The controller achieved a $1.45$~mm error for Target~1 or $1.7\%$ of the inserted needle length, but Target~2 (bending towards $u_x>0$) exhibited an $8.3$~mm error approximately $8\%$ of the inserted needle length. For the trajectory-tracking experiments the error remain below $6\%$ of the inserted needle length.

\section{Conclusion}\label{sec:conclusion}
This work presented the development and experimental validation of a bilinear model predictive control framework for steering the OncoReach tendon-driven robotic stylet compatible with clinically used brachytherapy needles.
Simulation studies demonstrated accurate convergence to fixed targets and effective tracking of complex three-dimensional trajectories. 
The proposed controller was subsequently validated through physical insertion experiments in tissue-mimicking phantom material using image-based needle tip tracking.
Experimental results confirmed the practical feasibility of the approach, demonstrating open-loop estimation errors below $3\%$ of the inserted length and closed-loop fixed-target tracking errors as low as $1.45$~mm or $1.7\%$ of the inserted length. However, the phantom tissue experiments also revealed that position errors can become relatively high in certain bending directions. 
Future work will focus on investigating these directional discrepancies by refining the calibration of the curvature-to-tension mapping and exploring potential twisting effects along the articulated tip. 
Furthermore, future efforts will aim at improving sensing accuracy through three-dimensional sensing modalities to overcome the limitations of 2D image-based measurements, reducing computational latency for faster control updates, and extending the framework to account for tissue deformation and model uncertainties during \emph{in vivo} procedures.

\bibliographystyle{IEEEtran}
\bibliography{references}

\end{document}